%% file: main.tex
\pdfoutput=1

\documentclass[11pt]{article}

\usepackage[preprint]{acl}

\usepackage{times}
\usepackage{latexsym}

\usepackage[T1]{fontenc}

\usepackage[utf8]{inputenc}

\usepackage{microtype}

\usepackage{inconsolata}

\usepackage{graphicx}

\usepackage{amsmath}
\usepackage{amsfonts}
\usepackage{enumitem}
\usepackage{subcaption}
\usepackage{booktabs}
\usepackage{tabularx}
\usepackage{multirow}
\usepackage{authblk}

\newcommand{\name}{A$^2$ATS}

\title{{\name}: Retrieval-Based KV Cache Reduction via Windowed Rotary Position Embedding and Query-Aware Vector Quantization}

\author{
    \textbf{Junhui He\textsuperscript{1,2 *}},
    \textbf{Junna Xing\textsuperscript{2 *\dag}},
    \textbf{Nan Wang\textsuperscript{2 *}},
    \textbf{Rui Xu\textsuperscript{2,3 *}},
    \textbf{Shangyu Wu$^\textbf{4}$},
    \\
    \textbf{Peng Zhou\textsuperscript{2}},
    \textbf{Qiang Liu\textsuperscript{2}},
    \textbf{Chun Jason Xue\textsuperscript{5}},
    \textbf{Qingan Li\textsuperscript{1 \dag}}
    \\
    \textsuperscript{1}Wuhan University,
    \textsuperscript{2}Alibaba Cloud Computing,
    \\
    \textsuperscript{3}Jinan University,
    \textsuperscript{4}City University of Hong Kong,
    \textsuperscript{5}MBZUAI
}

\begin{document}
\maketitle

\footnotetext{\textsuperscript{*}Equal contributions.}
\footnotetext{\textsuperscript{\dag}Corresponding authors.}

\begin{abstract}
\input{sections/0-abstract}
\end{abstract}

\input{sections/1-introduction}
\input{sections/3-background}
\input{sections/4-motivations}
\input{sections/5-methods}
\input{sections/6-experiments}
\input{sections/2-related_work}
\input{sections/7-conclusion}
\input{sections/8-limitations}
\input{sections/9-acknowledgements}

\bibliography{sections/references}

\clearpage
\appendix
\input{sections/appendix}

\end{document}

%% file: sections/0-abstract.tex
Long context large language models (LLMs) pose significant challenges for efficient serving due to the large memory footprint and high access overhead of KV cache.
Retrieval-based KV cache reduction methods can mitigate these challenges, typically by offloading the complete KV cache to CPU and retrieving necessary tokens on demand during inference.
However, these methods still suffer from unsatisfactory accuracy degradation and extra retrieval overhead.
To address these limitations, this paper proposes {\name}, a novel retrieval-based KV cache reduction method.
{\name} aims to obtain an accurate approximation of attention scores by applying the vector quantization technique to key states, thereby enabling efficient and precise retrieval of the top-K tokens.
First, 
we propose Windowed Rotary Position Embedding, which decouples the positional dependency from query and key states after position embedding.
Then, 
we propose query-aware vector quantization that optimizes the objective of attention score approximation directly.
Finally, 
we design the heterogeneous inference architecture for KV cache offloading, enabling long context serving with larger batch sizes.
Experimental results demonstrate that {\name} can achieve a lower performance degradation with similar or lower overhead compared to existing methods, thereby increasing long context serving throughput by up to $2.7 \times$.

%% file: sections/1-introduction.tex
\section{Introduction}

Large language models (LLMs) with long context windows~\citep{gpt-4, gemini-1.5, llama-3, mixtral, qwen-2.5, deepseek-v3} are driving advancements in AI applications. 
However, these models pose significant challenges for efficient serving. 
Their Transformer-based~\citep{transformer} architecture generates and maintains a Key-Value (KV) cache during inference to store intermediate results and avoid re-computation.  
As the context length increases, the size of the KV cache grows proportionally, leading to severe overheads.
First, the size of KV cache accessed when generating each token increases, resulting in a GPU memory bandwidth bottleneck. 
Moreover, the large KV cache size of each request limits the maximum feasible batch size, resulting in suboptimal GPU utilization.

Various methods were proposed to address these challenges from different perspectives.
Quantization-based methods~\citep{kivi, kvquant} compress KV cache by using lower bit-width representations for KV cache elements.
Eviction-based methods~\citep{streaming-llm, h2o, snapkv, pyramidinfer} reduce the KV cache size by directly evicting unimportant tokens from memory.
Retrieval-based methods~\citep{quest, loki, pqcache, clusterkv, magicpig} offload the complete KV cache to CPU memory and retrieve necessary tokens on demand during inference.
However, these methods still face challenges of limited compression ratio, unsatisfactory accuracy degradation, or extra retrieval overhead.

To address the above limitations, this paper proposes {\name}, a novel retrieval-based KV cache reduction method.
{\name} aims to obtain an \textbf{A}ccurate \textbf{A}pproximation of \textbf{AT}tention \textbf{S}cores by applying vector quantization technique to key states, thereby enabling efficient and precise retrieval of the top-K tokens.
In order to achieve this goal, we face two main challenges.
First, the position-dependent nature of key states after applying position embedding hinders the direct application of shared codebooks across varying inputs.
Second, directly utilizing the conventional vector quantization fails to guarantee an accurate approximation of attention scores.
To overcome these challenges, the main contributions in this paper are as follows:
\begin{itemize}[nosep]
    \item We observe high inter-input similarities between
    codebooks 
    of key states before position embedding,
    and the objective misalignment between vector quantization and attention score approximation by experimental and theoretical analysis;
    \item We propose Windowed Rotary Position Embedding to decouple the positional dependency from query and key states after position embedding,
    \item 
    We propose 
    and query-aware vector quantization that directly optimizes the objective of attention score approximation;
    \item We design the heterogeneous inference system for KV cache offloading, enabling long context serving with larger batch sizes.
\end{itemize}
Experimental results demonstrate that {\name} can achieve a low accuracy degradation of 2.2 on Llama-3.1-8B and 0.4 on Mistral-7B while accessing only 6\% of the entire KV cache, thereby increasing long context serving throughput by up to $2.7 \times$.
Our source code is publicly available\textsuperscript{1}.

\footnote{\textsuperscript{1}\url{https://github.com/junhuihe-hjh/A2ATS}}

%% file: sections/3-background.tex
\section{Preliminaries}



\subsection{Self-Attention Modules and Rotary Position Embedding}

\label{sec:rope}

Self-attention modules~\citep{transformer} and Rotary Position Embedding (RoPE)~\citep{rope} have become the de facto standard components of state-of-the-art (SOTA) LLMs~\citep{llama-3, qwen-2.5, mixtral, deepseek-v3}.

In the self-attention module, during decoding phase, the inference process begins by linearly projecting the input states of the $i$-th token into query ($q_i$), key ($k_i$), and value ($v_i$) states,  where $q_i, k_i, v_i \in \mathbb{R}^{1 \times d}$, and $d$ denotes the number of channels or hidden dimensions per head. 
To enable the model to effectively capture the positional relationships between tokens, position embeddings are then applied to the query and key states. 
These hidden states before and after this transformation are abbreviated as pre-PE and post-PE states, respectively.

RoPE is a commonly used position embedding in SOTA LLMs.
Specifically, for the $i$-th token, a position-dependent rotation matrix $R_i \in \mathbb{R}^{d \times d}$ is applied to the query $q_i$ and key $k_i$, to obtain their post-PE counterparts, denoted by $\tilde q_i$ and $\tilde k_i$:

\begin{equation}
    \tilde q_i = q_i R_i, \quad \tilde k_i = k_i R_i
\end{equation}
Then the matrices of KV cache of the context can be denoted by ${\tilde K} = [{\tilde k}_1; {\tilde k}_2; \dots; {\tilde k}_n] \in \mathbb R^{n \times d}$ and $V = [ v_1;  v_2; \dots;  v_n] \in \mathbb R^{n \times d}$ respectively, where $n$ denotes the context length.
Next, these post-PE states are used to compute the output state $o_i$ as shown in formula (\ref{formula:output state}):
\begin{equation}
    \label{formula:output state}
    \begin{aligned}
        o_i & = \operatorname{Softmax}\left( \frac{\tilde q_i {\tilde K}^\top}{\sqrt d} \right) V 
        = \operatorname{Softmax} \left( \frac{u_i}{\sqrt d} \right) V
    \end{aligned}
\end{equation}
where $u_i = \tilde q_i {\tilde K}^\top \in \mathbb R^{1 \times n}$ denotes the attention scores before softmax. 

Due to the inherent property of rotation matrices that $R_i R_j^\top = R_{i-j}$~\citep{rope}, the attention score $u_{i,j}$ between the $i$-th query and $j$-th key can be expressed as:
\begin{equation}
    \label{eq:rope}
    \begin{aligned}
        u_{i,j} = \tilde q_i {\tilde k}^\top_j = q_i R_i (k_j R_j)^\top 
        &= q_i R_i R_j^\top k_j^\top \\
        &= q_i R_{i-j} k_j^\top
    \end{aligned}
\end{equation}
This equation illustrates how RoPE encodes the relative position ($i-j$) directly into the attention scores, 
allowing the model to effectively capture
the positional relationships between tokens.

\subsection{Vector Quantization for Efficient Attention Score Approximation}

Vector quantization~\citep{vector-quantization} is a data compression technique that maps input vectors to a finite set of codewords from a learned codebook.

Formally, given an input space $X \subseteq \mathbb{R}^{1 \times d}$ with data distribution $\mathcal D$,
vector quantization aims to construct a codebook $C = \{c_1, c_2, \dots, c_L\} \subset \mathbb{R}^{1 \times d}$ with a size of $L$ codewords to minimize the following objective:
\begin{equation}
    \label{eq:vq_objective}
    J(C) = \mathbb E_{x \sim \mathcal D}[ \| x - \hat x \|^2 ]
\end{equation}
where $x \in \mathbb R^{1 \times d}$ denotes the input vector, $\hat x = c_{f(x; C)}$ denotes the quantized vector, and $f(x; C)$ denotes the quantization function that maps $x$ to its nearest codeword:
\begin{equation}
    f(x; C) = \operatorname*{argmin}_j \| x - c_j \|^2
\end{equation} 

Finding the optimal codebook $C$ is computationally expensive.
Therefore, approximate algorithms such as LBG and k-means++~\citep{lbg, kmeans++} are commonly used to find a suboptimal but effective codebook.

After obtaining the codebook, vector quantization compresses an input $x$ by replacing the original vector with its index $s = f(x; C)$.
Since the storage requirement for the index is substantially lower than that of the original vector, vector quantization achieves significant data compression ratio.

Multiple studies~\citep{transformer-vq, pqcache, clusterkv} have investigated applying vector quantization to post-PE key states of LLMs to efficiently approximate attention scores.
Let $s \in \{1, 2, \dots, L\}^{1 \times n}$ denotes the codeword index vector of all post-PE key states, where the length of this vector is $n$, and each element $s_i \in \{ 1, 2, \dots, L \}$ denotes the codeword index of the $i$-th key state.
Then, the $\tilde k_i$ can be quantized as $\hat {k}_i = c_{s_i}$
, and the attention score $u_{i,j}$ can be approximated as:
\begin{align}
    \label{eq:attention_weights_approximation}
    \hat u_{i,j} = \tilde q_i \hat k_j^\top = \tilde q_i c^\top_{s_j}
\end{align}
This equation illustrates the approximation of attention scores without the memory-intensive access to the $\tilde k_j$.

%% file: sections/4-motivations.tex
\begin{figure}[t]
    \centering
    \includegraphics[width=0.7\linewidth]{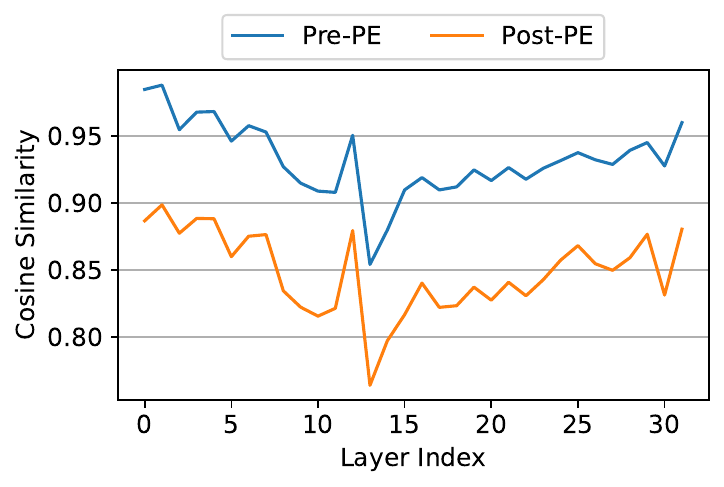}
    \caption{Inter-sample cosine similarities of pre-PE and post-PE codebooks.}
    \label{fig:cosine_similarity}
\end{figure}

\begin{figure}[t]
    \centering
    \begin{tabular}{cc}
        \begin{subfigure}[b]{0.225\textwidth}
            \includegraphics[width=\textwidth]{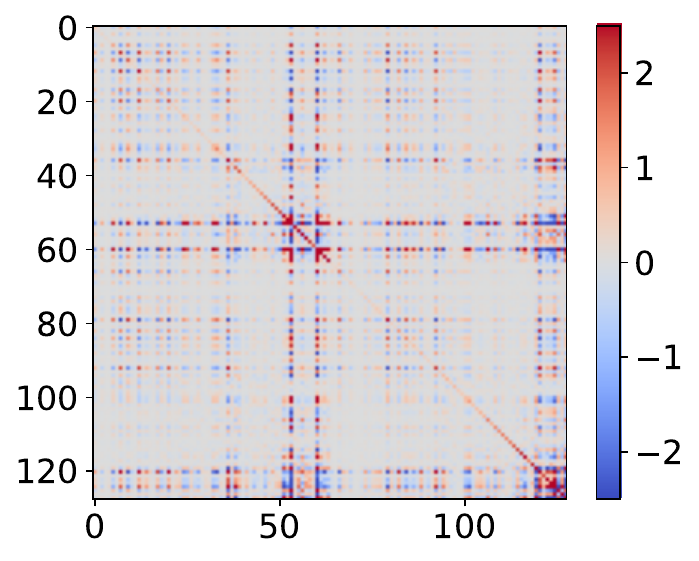}
            \caption{Layer 16, Head 8}
        \end{subfigure}
        &
        \begin{subfigure}[b]{0.225\textwidth}
            \includegraphics[width=\textwidth]{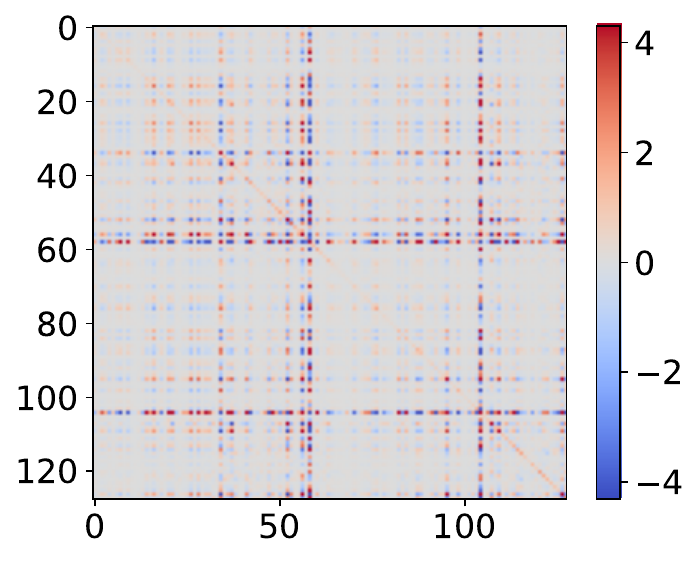}
            \caption{Layer 32, Head 8}
        \end{subfigure}
    \end{tabular}

    \caption{Visualization of second-moment matrices $H$ of post-PE query states. Each pixel represents an element in $H$. Warmer colors correspond to higher values, while cooler colors correspond to lower values.}

    \label{fig:hessian_matrix}
\end{figure}

\begin{figure}[t]
    \centering
    \includegraphics[width=0.7\linewidth]{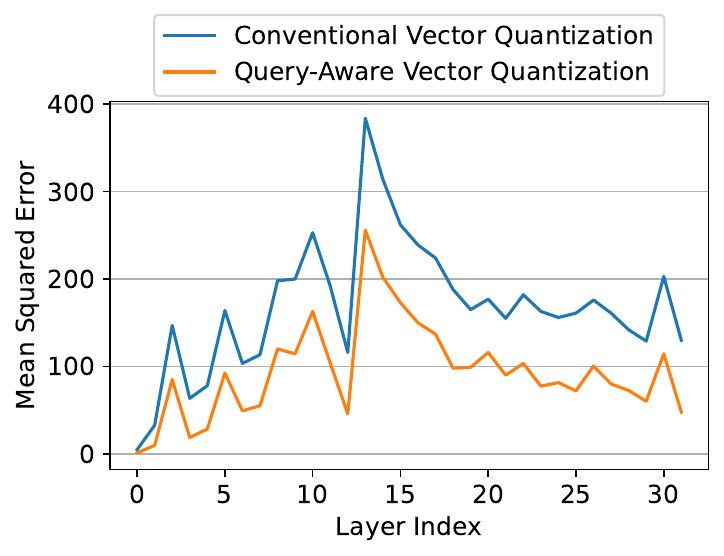}
    \caption{A comparison of MSE of attention score approximation between conventional vector quantization and query-aware vector quantization.}
    \label{fig:attention_score_mse}
\end{figure}

\section{Motivation}

\subsection{Inter-Input Similarity of Codebooks}

\label{sec:codebook_sharing}

PQCache~\citep{pqcache} and ClusterKV~\citep{clusterkv} propose applying vector quantization to post-PE key states, with individual codebooks constructed for each input during the prefilling phase.
However, constructing codebooks during inference requires iterative access to the entire KV cache, incurring high memory access overhead.
To address this limitation, we investigate the feasibility of employing shared codebooks for all inputs based on the inter-input similarity of codebooks.

To quantify the similarity between codebooks $C_1$ and $C_2$, we define the cosine similarity between codebooks as metric, which can be formulated as:
\begin{equation}
    \label{eq:codebook_similarity}
    \begin{aligned}
        \operatorname*{sim}(C_1, C_2) = & \frac 1 {2L} \sum_{i=1}^L \max_{j \in \{1, 2, \dots, L\}} \cos ({c_1}_i, {c_2}_j) + \\
        & \frac 1 {2L}  \sum_{i=1}^L \max_{j \in \{1, 2, \dots, L\}} \cos ({c_2}_i, {c_1}_j)
    \end{aligned}
\end{equation}
where ${c_1}_i,{c_1}_j \in C_1, {c_2}_i,{c_2}_j \in C_2$.
This metric computes the average maximum cosine similarity from each codeword in one codebook to any codeword in the other codebook.
A cosine similarity score closer to $1$ indicates higher similarity between codebooks, while a score closer to $0$ or negative values indicates lower similarity.

We utilize the Llama-3.1-8B-Instruct model~\citep{llama-3}, and two random samples from the FineWeb dataset~\citep{fineweb} with a context length of approximately 32k tokens each for experiments.
We collect pre-PE and post-PE (RoPE is used in here) key states from all attention heads on both samples, then employ k-means++ algorithm~\citep{kmeans++} to generate codebooks with a size of 4096 codewords for each set of key states, and calculate the inter-sample cosine similarities of pre-PE and post-PE codebooks using Equation~\ref{eq:codebook_similarity}. 
The similarities obtained are shown in  Figure~\ref{fig:cosine_similarity}.
From the experimental results, we derive the following key observations:

\noindent
\textbf{
    Observation 1: High inter-input similarities of pre-PE codebooks suggest the potential for using shared codebooks to effectively approximate key states across various inputs.
} 
The cosine similarities for pre-PE codebooks remain remarkably high, exceeding 0.9 for the majority of layers, indicate a very strong similarity across codebooks of various inputs. 
This finding indicates that the semantic information in key states might be broadly similar.

\noindent
\textbf{
    Observation 2: The position-dependent nature of post-PE key states hinders the direct application of a shared codebook.
} 
The post-PE codebook similarities are consistently lower, fluctuating around 0.85, with some layers below 0.8, indicating weaker similarity across inputs. 
The reason is that RoPE causes semantically similar key states at different positions to have different representations.
The position-dependent nature of post-PE key states makes it difficult to construct a single codebook that can effectively quantize representations with all semantic information at all possible positions.
This representation divergence becomes a key challenge for directly applying a shared codebook on post-PE key states, necessitating the development of a novel vector-quantization-compatible position embedding method.

\subsection{Objective Misalignment of Vector Quantization}

\label{sec:objective_mismatch}

The optimization objective of vector quantization is to minimize the mean squared error (MSE) of the approximate key states, while attention score approximation focuses on minimizing the MSE of attention scores.
This discrepancy between their optimization objectives raises a fundamental question:

\textit{Does the optimal codebook for vector quantization necessarily yield the most accurate approximation of attention scores?}

Formally, let $C$ denote the codebook constructed by vector quantization, $ J(C) $ denote the MSE of vector quantization with $C$, and $ J'(C) $ denote the MSE of attention scores approximation with $C$.
We need to investigate whether the following equation holds:
\begin{equation}
    \operatorname*{argmin}_C J(C) \equiv \operatorname*{argmin}_C J'(C)
\end{equation}
To address this question, we analyze the relationship between $ J $ and $ J' $.

For vector quantization, the objective $J$ can be reformulated as:
\begin{equation}
    \begin{aligned}
        J(C) & = \mathbb E_{\tilde k \sim \mathcal D^{\mathrm{key}}} [ \| \tilde k - \hat k \|^2 ] \\
        & = \mathbb E_{\tilde k \sim \mathcal D^{\mathrm{key}}} [ ( \tilde k - \hat k)( \tilde k - \hat k)^\top ]
    \end{aligned}
\end{equation}
where $\mathcal D^{\mathrm{key}}$ denotes the distribution of post-PE key states, and $\hat k$ denotes quantized key.

For attention score approximation, the objective $J'$ is formualted as:
\begin{equation}
    \label{eq:objective_attention_score_approximation}
    \begin{aligned}
        J'(C) & = \mathbb E_{\tilde k \sim \mathcal D^\mathrm{key},\ \tilde q \sim \mathcal D^\mathrm{query}}[(\tilde q \tilde k^\top - \tilde q \hat k^\top)^2] \\
        & = \mathbb E_{\tilde k \sim \mathcal D^\mathrm{key},\ \tilde q \sim \mathcal D^\mathrm{query}}[(\tilde q (\tilde k - \hat k)^\top)^2] \\
        & = \mathbb E_{\tilde k \sim \mathcal D^\mathrm{key},\ \tilde q \sim \mathcal D^\mathrm{query}}[(\tilde k - \hat k) \tilde q^\top \tilde q (\tilde k - \hat k)^\top] \\
        &= \mathbb E_{\tilde k \sim \mathcal D^\mathrm{key}} [(\tilde k - \hat k) H (\tilde k - \hat k)^\top]
    \end{aligned}
\end{equation}
where $\mathcal D^\mathrm{query}$ denotes the distribution of post-PE query states, 
$H = \mathbb E_{\tilde q \sim \mathcal D^\mathrm{query}}[\tilde q^\top \tilde q] \in \mathbb R^{d \times d}$ denotes the second-moment matrix of query states.

The two objectives only align if the query-dependent $H$ is proportional to the identity matrix.
To examine this consistency, we visualize $H$ on a set of input samples, as depicted in Figure~\ref{fig:hessian_matrix}.
The visualization shows that $ H $ is not proportional to the identity matrix, displaying a non-uniform and non-diagonal structure.
This result reveals a fundamental misalignment between the vector quantization objective $J$ and the attention score approximation objective $J'$, suggesting potential inaccuracy in attention score approximation.

To validate the impact of this objective mismatch, we compare conventional vector quantization that minimizes objective $J$, with its query-aware variant that directly minimizes objective $J'$ (implementation details in Section~\ref{sec:query_aware_vq}) in Figure~\ref{fig:attention_score_mse}. 
From experimental results, the query-aware method consistently achieves lower squared error in attention score approximation.
Thus, we derive the following observation:

\noindent
\textbf{
    Observservation 3:
    Optimizing vector quantization alone fails to guarantee accurate apprroximation of attention scores due to objective misalignment, necessitating query-aware vector quantization for bridging this objective gap. 
}

%% file: sections/5-methods.tex
\section{Method}

In {\name}, there are two stages.
During the offline pre-processing stage (1st stage), {\name} constructs a shared codebook on a representative dataset for each attention head of each layer.

During the inference stage (2nd stage), {\name} applies quantization functions to key states to map them to the nearest codewords.
At each autoregressive decoding step, {\name} first utilizes codebooks and codeword indices to approximate attention scores, then retrieves the top-K tokens with the highest attention scores for computation, thereby mitigating the memory overhead of accessing the entire KV cache.

\subsection{Windowed Rotary Position Embedding}

As discussed in Section~\ref{sec:codebook_sharing}, the position-dependent nature of post-PE key states hinders the direct application of a shared codebook in the vector quantization process.
A seemingly straightforward solution would be to quantize pre-PE key states, and then incorporate RoPE when approximating attention scores.
However, this approach is computationally expensive, as it necessitates calculating and applying the rotary matrices for each token at each inference.

To overcome this inefficiency, while eliminating the inherent position-dependent nature of post-PE key states, 
we propose Windowed Rotary Position Embedding (WRoPE).
This approach builds on the findings by~\citet{rerope, rerope-blog} that transformer-based models are nonsensitive to the positional information of non-local tokens.
The core idea of WRoPE is to use standard RoPE for local tokens (i.e. those in the window) and use approximate positional information for non-local tokens (i.e. those not in the window).
Specifically, WRoPE computes the attention scores as follows:
\begin{equation}
    \label{eq:wrope}
    u_{i,j} = 
    \begin{cases}
        \begin{aligned}
            q_i R_{i-j} k_j^\top, \quad & i-j < w \\
            q_i R_b k_j^\top, \quad & i-j \ge w
        \end{aligned}
    \end{cases}
\end{equation}
where $w$ is the window size, acting as a threshold for local vs. non-local tokens, and $b$ is a constant value representing a fixed relative position approximation for non-local tokens.

For local tokens (i.e., $i - j < w$), WRoPE functions identically to standard RoPE, as defined in Equation~\ref{eq:rope}.
For non-local tokens outside the window (i.e., $i - j \ge w$), the position-dependent rotation matrix $R_{i-j}$ is replaced by a fixed rotation matrix $R_b$, approximating the relative positional information $ (i-j) $ with a constant offset $b$.
Then, we can calculate the post-PE query and key states as:
\begin{equation}
    \tilde q_i = q_iR_b, \quad \tilde k_i = k_i
\end{equation}

Since post-PE key $\tilde k_i$ states are identical to their pre-PE counterparts $k_i$, WRoPE decouples the positional dependency from post-PE representations, therefore optimizes subsequent vector quantization.

\subsection{Query-Aware Vector Quantization}

\label{sec:query_aware_vq}

As discussed in Section \ref{sec:objective_mismatch}, conventional vector quantization fails to achieve accurate approximation of attention scores, due to the objective misalignment between vector quantization and attention score approximation. 

To address this limitation, \textbf{we propose query-aware vector quantization, a custom vector quantization method that directly optimizes the objective of attention score approximation. }
Specifically, we replace the squared Euclidean distance $\|\tilde k - \hat k\|^2$ of conventional vector quantization with a query-aware quadratic form $(\tilde k - \hat k) H (\tilde k - \hat k)^\top$ derived from formula (\ref{eq:objective_attention_score_approximation}), where $H$ represents the second-moment matrix of query states. 

Formally, the query-aware vector quantization minimizes the following objective:
\begin{equation}
    \label{eq:query_aware_vq}
    J'(C) = \mathbb E_{\tilde k \sim \mathcal {D^\mathrm{key}}} \left[(\tilde k - \hat k) H (\tilde k - \hat k)^\top\right]
\end{equation}
where $\hat k = c_{f'(\tilde k; C)}$ denotes the quantized $\tilde k$, and the corresponding query-aware quantization vector quantization is formulated as:
\begin{equation}
    f'(\tilde k; C) = \operatorname*{argmin}_j (\tilde k - c_j) H (\tilde k - c_j)^\top    
\end{equation}

For the codebook construction process, we reformulate the objective function to utilize conventional efficient vector quantization algorithms like k-means++~\citep{kmeans++}.
Specifically, we apply Cholesky decomposition to the positive definite matrix $H = LL^T$, where $L \in \mathbb R^{d \times d}$ denotes the Cholesky factor.
Let
\begin{equation}
    \label{eq:definition_z}
    z = \tilde k L, \quad C^z = CL, \quad\hat z = \hat k L
\end{equation}
where $z \in \mathbb R^{1 \times d}$ denotes the transformed key state.
Then, we can re-derive the objective of attention score approximation $J'$ as:
\begin{equation}
  \label{eq:objective_transform}
    \begin{aligned}
        J'(C) & = \mathbb E_{\tilde k \sim \mathcal {D^\mathrm{key}}} \left[(\tilde k - \hat k)H(\tilde k - \hat k)^\top\right] \\
        & = \mathbb E_{\tilde k \sim \mathcal {D^\mathrm{key}}} \left[(\tilde k - \hat k)LL^\top(\tilde k - \hat k)^\top\right] \\
        & = \mathbb E_{\tilde k \sim \mathcal {D^\mathrm{key}}} \left[(\tilde kL - \hat kL)(\tilde kL - \hat kL)^\top\right] \\
        & = \mathbb E_{z \sim D^z} \left[(z - \hat z)(z - \hat z)^\top\right]
    \end{aligned}
\end{equation}
And the quantization function $f'$ can be re-derived as:
\begin{equation}
    \begin{aligned}
        f'(\tilde k; C) & = \operatorname*{argmin}_j (\tilde k - c_j) H (\tilde k - c_j)^\top     \\
        & = \operatorname*{argmin}_j (\tilde k L - c_jL)(\tilde k L - c_jL)^\top \\
        & = \operatorname*{argmin}_j (z - c^z_j)(z - c^z_j)^\top \\
        & = f(z; C^z)
    \end{aligned}
\end{equation}
where $f(z; C^z)$ denotes the quantization function of conventional vector quantization.
Then, we can derive that
\begin{equation}
    \label{eq:quantization_function_transform}
    \begin{aligned}
        \hat z = \hat k L & = c_{f'(\tilde k; C)}L = c^z_{f(z; C^z)} 
    \end{aligned}
\end{equation}

Equations~\ref{eq:objective_transform},~\ref{eq:quantization_function_transform}
reveal that \textbf{the objective of attention score approximation is equivilant to that of conventional vector quantization on transformed $z$}.
This alignment enables the application of conventional efficient vector quantization algorithms in codebook construction process.

During the offline pre-processing stage, we collect $z$ on a representative dataset and construct its codebook $C^z$ using k-means++~\citep{kmeans++}.
Then, The original shared codebook $C$ for $\tilde k$ is calculated as:
\begin{equation}
    C = C^z L^{-1}
\end{equation}
During inference, the codeword index of $\tilde k$ is computed through query-aware quantization function:
\begin{equation}
    f'(\tilde k; C) = \operatorname*{argmin}_j (\tilde k L - c_jL)(\tilde k L - c_jL)^\top
\end{equation}
Let $s \in \{1, 2, \dots, L\}^{1 \times n}$ denotes the codeword index vector of all key states after applying query-aware vector quantization, where $s_j = f'(\tilde k_j; C)$.
Then, the attention score is approximated as:
\begin{equation}
    \hat u_{i,j} = \tilde q_i \hat k_j = \tilde q_i c_{s_j}
\end{equation}

\subsection{Heterogeneous Inference Design}

Although approximating attention scores via vector quantization and then selectively retrieving top-K tokens for computation reduces memory access overhead, the issue of KV Cache occupying substantial GPU memory remains unresolved. 
To reduce the memory footprint of KV Cache and enable larger batch sizes for improved GPU utilization, we design a heterogeneous inference system. 

We partition the decoding process of our proposed {\name} into three components:  

\noindent (1) \textbf{GPU-based model execution}: All model weights reside on the GPU memory. 
Computations involving model weights are executed on the GPU during inference.  

\noindent(2) \textbf{GPU-based approximation of attention scores}: The codebook is stored on the GPU. 
During inference, the GPU first executes the quantization function to assign codewords to key states, then computes attention weight approximations using the codebooks and indices, and lastly gathers the indices of top-K tokens.

\noindent(3) \textbf{CPU-based selective attention}: The full KV Cache is maintained on the CPU memory. 
During decoding, the top-K token indices and the current query state are transferred to the CPU, where selective attention computation is performed to derive the attention output. 
This output is then transferred back to the GPU for subsequent computations.

This design aims to minimize data transfer between CPU and GPU, thereby reducing latency. 
Furthermore, to fully leverage the CPU's thread-level parallelism and SIMD capabilities, we implement a custom selective attention kernel optimized for CPU execution.

%% file: sections/6-experiments.tex
\section{Experiments}

\subsection{Experimental Setup}

\label{sec:experimental-setup}

\noindent \textbf{Tasks.} 
We utilize RULER~\citep{ruler} as our benchmark for downstream tasks evaluation.
This synthetic benchmark contains thirteen subtasks organized into four categories: information retrieval, multi-hop tracing, information aggregation, and question answering. 
It evaluates long-context comprehension and reasoning capabilities of LLMs, while effectively revealing the accuracy drop caused by KV cache reduction methods.

\noindent \textbf{Models.}
We conduct our main experiments on Llama-3.1-8B-Instruct~\citep{llama-3} and MegaBeam-Mistral-7B-512k~\citep{megabeam-mistral}.
These models feature long-context processing capabilities with context windows of up to 128K and 512K tokens, respectively. 
As for the ablation study and end-to-end throughput evaluation, we apply the Llama-3.1-8B-Instruct model.

\noindent \textbf{Methods.}
For main experiments, we compare the proposed {\name} with the following four KV cache reduction methods, along with the full attention baseline: H2O~\citep{h2o}, SnapKV~\citep{snapkv}, Quest~\citep{quest}, MagicPIG~\citep{magicpig}.
For a fair comparison, the sparsity ratios of all KV cache reduction methods are controlled around 0.06.
Detailed discussions are presented in Appendix~\ref{appendix:baselines}.

For ablation studies, we evaluate the following configurations on Llama-3.1-8B-Instruct:
\begin{itemize}[nosep]
    \item \textbf{Baseline}: It utilizes standard RoPE and conventional vector quantization.
    \item \textbf{WRoPE}: 
    It utilizes WRoPE and conventional vector quantization.
    \item \textbf{QAVQ}:
    It utilizes standard RoPE and query-aware vector quantization.
    \item \textbf{{\name}}: The proposed method with WRoPE and query-aware vector quantization.
\end{itemize}

\noindent \textbf{Implementation Details.}
The hyper-parameters $w$ and $b$ of WRoPE are set to $64$ and $2048$, respectively.
The codebooks of query-aware vector quantization, with a size of 4096 each, are constructed from a set of sample inputs consisting of approximately 64K tokens of text randomly sampled from FineWeb~\citep{fineweb} and 16K tokens of randomly generated uuid strings.
The end-to-end speedup experiments are conducted on a server equipped with an NVIDIA H800 GPU with 80GB memory, and an Intel Xeon Platinum 8469C CPU.

\subsection{Main Results on Downstream Tasks}

\begin{table*}[htb]
    \centering
    \resizebox{0.8\textwidth}{!}{
        \begin{tabular}{l|cc|cccc|c}
            \toprule   
    		\multirow{2}*{\textbf{Models}} &
            \multirow{2}*{\textbf{Sparsity$\downarrow$}} & 
            \multirow{2}*{\textbf{Aux Mem$\downarrow$}}& 
            \multicolumn{5}{c}{\textbf{Accuracy$\uparrow$}} \\
                
            \cmidrule{4-8}
        
            ~ & ~ & ~ &
            \textbf{16K} & 
            \textbf{32K} & 
            \textbf{64K}  & 
            \textbf{96K}  & 
            \textbf{Average} \\
            
            \midrule

            \textit{Llama-3.1-8B-Instruct} &
            1.000 &
            0.000 &
            94.4 &
            91.9 &
            85.9 &
            83.1 &
            88.8 \\

            H2O &
            0.060 &
            0.008 &
            27.6 &
            30.6 &
            24.9 &
            25.0 &
            27.0 \\

            SnapKV &
            0.060 &
            0.008 &
            72.7 &
            75.1 &
            72.2 &
            70.7 &
            72.7 \\

            Quest &
            0.060 &
            0.031 &
            84.3 &
            84.0 &
            80.0 &
            74.4 &
            80.7 \\

            MagicPIG &
            0.068 &
            2.344 &
            \textbf{92.3} &
            87.6 &
            83.9 &
            79.1 &
            85.7 \\

            {\name} &
            0.060 &
            0.008 &
            92.2 &
            \textbf{90.4} &
            \textbf{84.3} &
            \textbf{79.6} &
            \textbf{86.6} \\

            \midrule

            \textit{MegaBeam-Mistral-7B-512K} &
            1.000 &
            0.000 &
            91.8 &
            88.2 &
            83.3 &
            83.4 &
            86.7 \\

            H2O & 
            0.060 & 
            0.008 &
            22.5 &
            23.4 &
            20.7 &
            22.6 &
            22.3 \\

            SnapKV &
            0.060 &
            0.008 &
            69.3 &
            68.5 &
            69.5 &
            65.2 &
            67.6 \\

            Quest &
            0.060 &
            0.031 &
            81.5 &
            80.8 &
            76.7 &
            74.4 &
            78.4 \\

            MagicPIG &
            0.064 & 
            2.344 &
            88.7 &
            85.2 &
            82.6 &
            81.8 &
            84.6 \\

            {\name} &
            0.062 &
            0.008 &
            \textbf{91.6} &
            \textbf{88.1} &
            \textbf{83.4} &
            \textbf{82.2} &
            \textbf{86.3} \\
            
            \bottomrule
        \end{tabular}
    }
    \caption{Comparison of sparsity ratio, auxiliary memory usage and accuracy on RULER benchmark. `Aux Mem' refers to `Auxialiary Memory Usage', which denotes the extra memory usage caused by KV cache reduction methods compared to the original key cache. `16K', `32K', `64K' and `96K' denote the input context length.}
    \label{tab:main_experiments}
\end{table*}

Table~\ref{tab:main_experiments} compares the accuracies of different methods on RULER, along with attention sparsity ratios and auxiliary memory overhead.
Experimental results draw the following conclusions:

\noindent \textbf{{\name} minimizes accuracy degradation under comparable sparsity ratios.}
For Llama models,
{\name} achieves an average accuracy of 86.6, outperforming H2O (27.0), SnapKV (72.7), Quest (80.7), and MagicPIG (85.7).
For Mistral models,
{\name} achieves an average accuracy of 86.3, surpassing H2O (22.3), SnapKV (67.6), Quest (78.4), and MagicPIG (84.6).
Notably, {\name} achieves these accuracies with comparable sparsity ratios (0.060 for Llama, 0.062 for Mistral) to H2O, SnapKV and Quest, lower than those of MagicPIG (0.068 for Llama, 0.064 for Mistral). 
These results demonstrate the superior accuracy preservation of {\name}.

\noindent \textbf{{\name} causes comparable or lower auxiliary memory overhead compared to existing methods.}
{\name} causes auxiliary memory usage (0.008) identical to eviction-based methods, i.e., H2O and SnapKV, while being significantly more memory efficient than retrieval-based approaches, i.e., Quest (0.031) and MagicPIG (2.344).
This efficiency comes from the inherent nature of vector quantization, requiring only one codeword index for each token in each attention head, eliminating the need for storing either per-page metadata (Quest) or large LSH tables (MagicPIG).

\subsection{Ablation Study}

\begin{table}[htb]
    \centering
    \resizebox{1.0\columnwidth}{!}{
        \begin{tabular}{l|cccc|c}
        
            \toprule
            
            \textbf{Config} & 
            \textbf{16K$\uparrow$} & 
            \textbf{32K$\uparrow$} & 
            \textbf{64K$\uparrow$}  & 
            \textbf{96K$\uparrow$}  & 
            \textbf{Average$\uparrow$} \\
            
            \midrule

            Baseline &
            86.4 &
            86.3 &
            81.5 &
            71.3 &
            81.4 \\

            WRoPE &
            \textbf{92.3} &
            90.0 &
            82.8 &
            78.4 &
            85.9 \\

            QAVQ &
            91.7 &
            86.9 &
            76.3 &
            69.4 &
            81.1 \\
            
            {\name} &
            92.2 &
            \textbf{90.4} &
            \textbf{84.3} &
            \textbf{79.6} &
            \textbf{86.6} \\
            
            \bottomrule
        \end{tabular}
    }
    \caption{Ablation study on the importance of WRoPE and query-aware vector quantization for accuracy. 
    }
    \label{tab:ablation}
\end{table}

Table~\ref{tab:ablation} validates the effectiveness of WRoPE and query-aware vector quantization
on improving model accuracy.
Experimental results draw the following conclusions: 
(1) WRoPE is fundamental to attention score approximation using shared codebooks.
(2) Query-aware vector quantization provides a further improvement in model accuracy by aligning the objectives of vector quantization and attention score approximation.
Detailed analysis is presented in Appendix~\ref{appendix:ablation}.

\subsection{End-to-End Inference Speedup}

\begin{figure}[t]
    \centering
    \begin{tabular}{cc}
        \begin{subfigure}[b]{0.45\columnwidth}
            \centering
            \includegraphics[width=1.0\columnwidth]{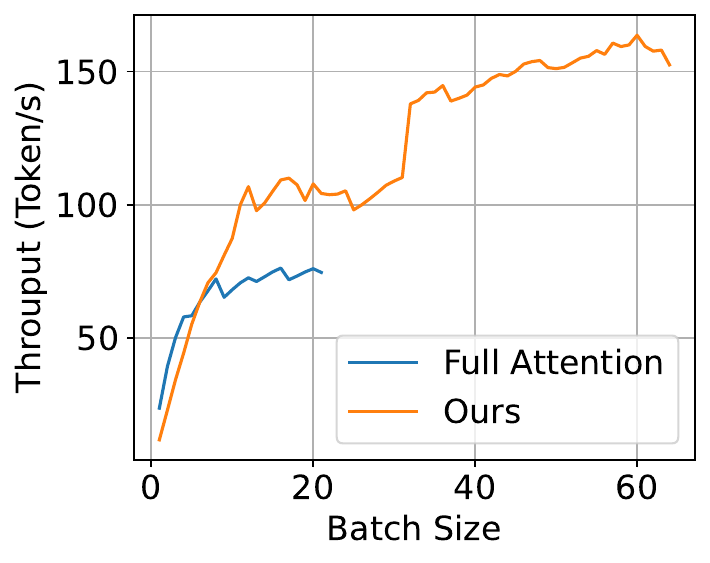}
            \caption{Inference throughput with a context length of 16K.}
        \end{subfigure}
        & 
        \begin{subfigure}[b]{0.45\columnwidth}
            \centering
            \includegraphics[width=1.0\columnwidth]{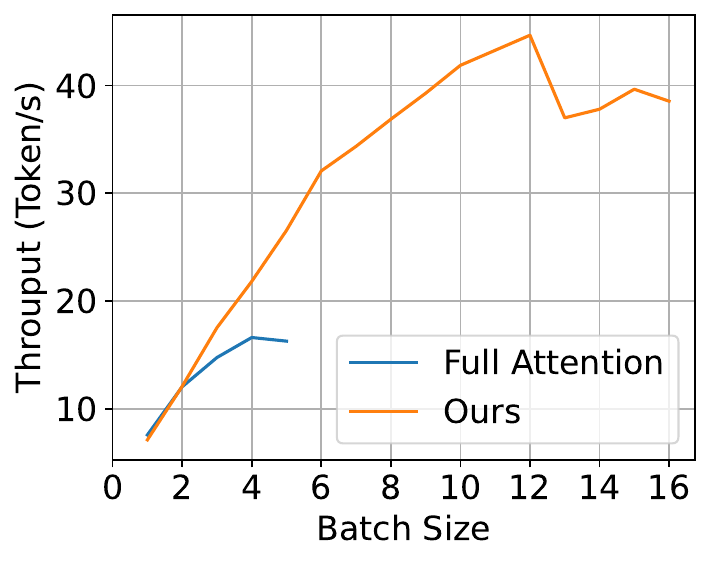}
            \caption{Inference throughput with a context length of 64K.}
        \end{subfigure}
    \end{tabular}

    \caption{End-to-end inference throughput of Llama-3.1-8B-Instruct across varying context lengths and batch sizes.}
    \label{fig:throughput}
\end{figure}

Figure~\ref{fig:throughput} compares the inference throughput of the proposed {\name} against full attention on Llama-3.1-8B-Instruct.
At a context length of 16K, {\name} initially exhibits marginally lower throughput than full attention for small batch sizes $(\leq 5)$, primarily due to CPU-GPU data transfer overhead.
As batch sizes increase, the throughput of {\name} grows linearly to exceed 100 tokens/s, while full attention plateaus below 80 tokens/s due to memory bandwidth bottleneck and ends up out of memory at a batch size of 22.
{\name} achieves a peak throughput of over 160 tokens/s, a 2.1× speedup over full attention, with a maximum batch size of over 64.
This trend becomes more pronounced at 64K context lengths, where full attention struggles with batches over 5, while {\name} serving a batch size of up to 16 with a throughput of up to 45 tokens/s, delivering a $2.7 \times$ performance advantage.
These results highlight {\name}'s potential in mitigating the memory bottleneck in long context LLM serving.

%% file: sections/2-related_work.tex
\section{Related Work}



\noindent\textbf{Quantization-based KV cache reduction.}
This method aims to compress KV cache by using lower bit-width representations for KV cache elements~\citep{kivi, kvquant}.
However, it typically faces challenges of limited compression ratio and extra computational overhead caused by the dequantization process.

\noindent\textbf{Eviction-based KV cache reduction.}
This method aims to reduce the KV cache size by directly evicting unimportant tokens from memory~\citep{streaming-llm, h2o, snapkv, pyramidinfer}. 
These methods typically record statistics of attention weights of each token. 
When the KV cache reaches its capacity limit, they utilize heuristic rules and historical statistics to predict which tokens are more likely to get high attention weights in future decoding, then retain these tokens while evicting the rest. 
Although these methods generally have low additional overhead, they often lead to noticeable performance degradation.

\noindent\textbf{Retrieval-based KV cache reduction.}
This method keeps the entire KV cache in memory while selectively retrieving tokens crucial for the current inference. 
Quest~\citep{quest} chunks the continuous KV cache into pages and pre-calculates necessary metadata for each page during prefilling. 
For decoding, it selects the top-K critical cache pages to participate in selective attention computation. 
PQCache~\citep{pqcache} and ClusterKV~\citep{clusterkv} perform vector quantization on the key states with individual codebooks constructed for each input during prefilling.
For decoding, the system uses codewords of the codebooks to approximate attention scores, then retrieves the top-K tokens for computation. 
MagicPIG~\citep{magicpig} employs Locality-Sensitive Hashing on query and key states for token retrieval.
Although these methods generally achieve lower performance degradation compared to eviction-based methods, they still suffer from unsatisfactory performance and lead to extra overhead.

%% file: sections/7-conclusion.tex
\section{Conclusion}
In this paper, we propose {\name}, a novel retrieval-based KV cache reduction method.
First, we propose Windowed Rotary Position Embedding to decouple the positional dependency from query and key states after position embedding.
Then, we propose query-aware vector quantization to achieve an accurate attention score approximation.
Next, we introduce the heterogeneous inference design for KV cache offloading, which increases available batch size.
Experimental results demonstrate that {\name} achieves lower performance degradation with comparable or lower overhead compared to existing methods, thereby boosting long context serving throughput by up to $2.7 \times$.

%% file: sections/8-limitations.tex
\section{Limitations}

The limitations of this work can be summarized in two main aspects.

First, while {\name} demonstrates lower accuracy degradation compared to existing methods while accessing a comparable proportion of KV cache,
it still exhibits non-negligible performance degradation.
This suggests opportunities for future work to investigate adaptive attention sparsity allocation strategies that dynamically optimize the sparsity ratios across layers and attention heads, based on their contextual importance.

Second, while {\name} increases long context serving throughput by up to $2.7 \times$, our current implementation is limited to single-GPU deployment.
Future research could further explore (1) distributed multi-GPU system designs for scaled deployment, (2) integration with disaggregated LLM serving architectures like MoonCake~\citep{mooncake}.

%% file: sections/9-acknowledgements.tex
\section*{Acknowledgements}

We thank all the reviewers for their insightful comments. This work is supported by the National Natural Science Foundation of China (No. 62472330).

%% file: sections/appendix.tex
\section{Method Configurations}

\label{appendix:baselines}

For the main experiments, we compare the following five KV cache reduction methods, along with the full attention baseline:

\begin{itemize}[nosep]
    \item \textbf{H2O}~\citep{h2o}: An eviction-based method that preserves heavy hitter tokens and recent tokens;
    \item \textbf{SnapKV}~\citep{snapkv}: An eviction-based method that preserves important tokens based on statistics within an observation window;
    \item \textbf{Quest}~\cite{quest}: A retrieval-based method that selects tokens based on metadata of KV cache pages;
    \item \textbf{MagicPIG}~\cite{magicpig}: A retrieval-based method that utilizes LSH for token sampling;
    \item \textbf{{\name}}: The proposed method.
\end{itemize}

For a fair comparison, the sparsity ratios of all KV cache reduction methods are controlled around 0.06, which means approximately 6\% of KV cache is accessed at each inference.
It is important to note that the sparsity ratio discussed in this paper differs in definition from the $\mathrm{cost}_2$ in MagicPIG~\citep{magicpig}. 
Specifically, $\mathrm{cost}_2$ measures the ratio of computation overhead (FLOPs) compared to full attention, whereas our sparsity ratio measures the ratio of memory access overhead (MOPs) relative to full attention. 
Since attention modules are typically considered memory-bound~\citep{transformer-survey}, we argue that the latter metric provides more meaningful insights on potential overhead reduction.
Additionally, the initial 4 tokens and the most recent 64 tokens are statically preserved to align with MagicPIG~\citep{streaming-llm}.
The detailed configurations for each methods are shown in Table~\ref{tab:configs}.

\begin{table}[ht!]
    \resizebox{\columnwidth}{!}{
        \begin{tabular}{c|c}
            \toprule
                
            \textbf{Method} & \textbf{Configurations} \\ 
            \midrule
            H2O 
            & \texttt{hh\_size} = 0.06 $\times$ \texttt{input\_length} \\
            
            SnapKV & 
            \texttt{prompt\_capacity} = 0.06 $\times$ \texttt{input\_length} \\
            
            Quest & 
            \texttt{page\_size} = 32, \texttt{ratio} = 0.06 \\
            
            MagicPIG & 
            \texttt{K} = 10, \texttt{L} = 150 \\
            
            {\name} & 
            \texttt{topk} = 0.03 \\
            
            \bottomrule
        \end{tabular}
    }
    \caption{Configurations of KV cache reduction methods.}
    \label{tab:configs}
\end{table}

\section{Detailed Ablation Study Analysis}

\label{appendix:ablation}

Table~\ref{tab:ablation} validates the effectiveness of WRoPE and query-aware vector quantization
on improving model accuracy.
Experimental results draw the following conclusions: 

\noindent \textbf{WRoPE is fundamental to attention score approximation using shared codebooks.}
WRoPE achieves an average improvement of +4.5 over the baseline, with consistent gains across all context lengths.
This result confirms WRoPE's critical role in preventing representation divergence of key states caused by positional embedding.

\noindent \textbf{Query-aware vector quantization provides a further improvement in model accuracy by aligning the objectives of vector quantization and attention score approximation}.
Our full method, incorporating query-aware vector quantization, demonstrates further improvements, particularly at longer context lengths (+1.5 at 64K, +1.2 at 96K, respectively).
However, query-aware vector quantization alone does not outperform conventional vector quantization.
This is because query-aware vector quantization relies on the estimation of the second-moment matrix of queries, i.e., $H = \mathbb E_{\tilde q \sim \mathcal D^{\text{query}}}[\tilde q^\top \tilde q]$ (Section~\ref{sec:query_aware_vq}, Equation~\ref{eq:query_aware_vq}). Positional dependencies induced by RoPE cause representation divergence in $\tilde q^\top \tilde q$, hindering the accurate estimation of $H$ and leading to decreased accuracy. The synergy between WRoPE (which eliminates positional dependencies) and query-aware quantization (which aligns the objective between vector quantization and attention score approximation) ultimately delivers state-of-the-art performance.

\section{Additional Experiments on LongBench}

To further validate the effectiveness of our proposed method on real-world tasks, we conducted additional experiments on 6 tasks from LongBench~\citep{longbench} (HotpotQA, MultiFieldQA-en, QMSum, TriviaQA, PassageRetrieval-en and RepoBench-P). 
The configurations of all KV cache reduction methods are the same as those presented Section~\ref{sec:experimental-setup}.

\begin{table*}[htb]
    \centering
    \resizebox{0.8\textwidth}{!}{
        \begin{tabular}{c|c|cccccc|c}
            \toprule
            \multirow{2}*{\textbf{Models}} &
            \multirow{2}*{\textbf{Sparsity$\downarrow$}} & 
            \multicolumn{7}{c}{\textbf{Accuracy$\uparrow$}} \\
            \cmidrule{3-9}
            ~ & ~ &
            \textbf{HpQA} & 
            \textbf{MfQA} & 
            \textbf{QMS}  & 
            \textbf{TrQA}  & 
            \textbf{PRe} &
            \textbf{RBP} &
            \textbf{Average} \\
            \midrule
            \textit{Llama-3.1-8B-Instruct} & 1.00 & 58.41 & 56.43 & 25.01 & 91.47 & 100.0 & 56.37 & 64.62 \\ 
            H2O & 0.060 & 57.44 & 49.72 & 24.09 & 91.55 & \textbf{99.50} & 52.28 & 62.43 \\ 
            SnapKV & 0.060 & 57.38 & 54.12 & 24.59 & 90.92 & \textbf{99.50} & 54.36 & 63.48 \\ 
            Quest & 0.060 & 57.79 & 55.49 & 24.58 & 90.70 & \textbf{99.50} & 53.94 & 63.67 \\ 
            MagicPIG & 0.068 & 57.80 & 55.99 & \textbf{25.26} & 90.82 & \textbf{99.50} & 55.31 & 64.11 \\ 
            {\name} & 0.060 & \textbf{58.03} & \textbf{56.84} & 25.03 & \textbf{91.63} & \textbf{99.50} & \textbf{56.27} & \textbf{64.55} \\ 
            \toprule
        \end{tabular}
    }
    \caption{Experimental results on LongBench.}
    \label{tab:longbench}
\end{table*}

Table~\ref{tab:longbench} compares the accuracies of different methods, along with attention sparsity ratios. 
Experimental results demonstrate that the proposed {\name} outperforms other baselines under comparable sparsity, emphasizing the effectiveness of our proposed method on a broader range of long-context tasks.

%% file: main.bbl
\begin{thebibliography}{31}
\providecommand{\natexlab}[1]{#1}

\bibitem[{Arthur and Vassilvitskii(2007)}]{kmeans++}
David Arthur and Sergei Vassilvitskii. 2007.
\newblock \href {http://dl.acm.org/citation.cfm?id=1283383.1283494} {k-means++: the advantages of careful seeding}.
\newblock In \emph{Proceedings of the Eighteenth Annual {ACM-SIAM} Symposium on Discrete Algorithms, {SODA} 2007, New Orleans, Louisiana, USA, January 7-9, 2007}, pages 1027--1035. {SIAM}.

\bibitem[{Bai et~al.(2024)Bai, Lv, Zhang, Lyu, Tang, Huang, Du, Liu, Zeng, Hou, Dong, Tang, and Li}]{longbench}
Yushi Bai, Xin Lv, Jiajie Zhang, Hongchang Lyu, Jiankai Tang, Zhidian Huang, Zhengxiao Du, Xiao Liu, Aohan Zeng, Lei Hou, Yuxiao Dong, Jie Tang, and Juanzi Li. 2024.
\newblock \href {https://doi.org/10.18653/v1/2024.acl-long.172} {{L}ong{B}ench: A bilingual, multitask benchmark for long context understanding}.
\newblock In \emph{Proceedings of the 62nd Annual Meeting of the Association for Computational Linguistics (Volume 1: Long Papers)}, pages 3119--3137, Bangkok, Thailand. Association for Computational Linguistics.

\bibitem[{Buzo et~al.(1980)Buzo, Jr., Gray, and Markel}]{vector-quantization}
Andres Buzo, Augustine H.~Gray Jr., Robert~M. Gray, and John~D. Markel. 1980.
\newblock \href {https://doi.org/10.1109/ICASSP.1980.1170996} {Speech coding based upon vector quantization}.
\newblock In \emph{{IEEE} International Conference on Acoustics, Speech, and Signal Processing, {ICASSP} '80, Denver, Colorado, USA, April 9-11, 1980}, pages 15--18. {IEEE}.

\bibitem[{Chen et~al.(2025)Chen, Sadhukhan, Ye, Zhou, Zhang, Nolte, Tian, Douze, Bottou, Jia, and Chen}]{magicpig}
Zhuoming Chen, Ranajoy Sadhukhan, Zihao Ye, Yang Zhou, Jianyu Zhang, Niklas Nolte, Yuandong Tian, Matthijs Douze, L{\'{e}}on Bottou, Zhihao Jia, and Beidi Chen. 2025.
\newblock \href {https://openreview.net/forum?id=ALzTQUgW8a} {Magicpig: {LSH} sampling for efficient {LLM} generation}.
\newblock In \emph{The Thirteenth International Conference on Learning Representations, {ICLR} 2025, Singapore, April 24-28, 2025}. OpenReview.net.

\bibitem[{DeepSeek{-}AI et~al.(2024)DeepSeek{-}AI, Liu, Feng, Xue, Wang, Wu, Lu, Zhao, Deng, Zhang, Ruan, Dai, Guo, Yang, Chen, Ji, Li, Lin, Dai, Luo, Hao, Chen, Li, Zhang, Bao, Xu, Wang, Zhang, Ding, Xin, Gao, Li, Qu, Cai, Liang, Guo, Ni, Li, Wang, Chen, Chen, Yuan, Qiu, Li, Song, Dong, Hu, Gao, Guan, Huang, Yu, Wang, Zhang, Xu, Xia, Zhao, Wang, Zhang, Li, Wang, Zhang, Zhang, Tang, Li, Tian, Huang, Wang, Zhang, Wang, Zhu, Chen, Du, Chen, Jin, Ge, Zhang, Pan, Wang, Xu, Zhang, Chen, Li, Lu, Zhou, Chen, Wu, Ye, Ye, Ma, Wang, Zhou, Yu, Zhou, Pan, Wang, Yun, Pei, Sun, Xiao, and Zeng}]{deepseek-v3}
DeepSeek{-}AI, Aixin Liu, Bei Feng, Bing Xue, Bingxuan Wang, Bochao Wu, Chengda Lu, Chenggang Zhao, Chengqi Deng, Chenyu Zhang, Chong Ruan, Damai Dai, Daya Guo, Dejian Yang, Deli Chen, Dongjie Ji, Erhang Li, Fangyun Lin, Fucong Dai, Fuli Luo, Guangbo Hao, Guanting Chen, Guowei Li, H.~Zhang, Han Bao, Hanwei Xu, Haocheng Wang, Haowei Zhang, Honghui Ding, Huajian Xin, Huazuo Gao, Hui Li, Hui Qu, J.~L. Cai, Jian Liang, Jianzhong Guo, Jiaqi Ni, Jiashi Li, Jiawei Wang, Jin Chen, Jingchang Chen, Jingyang Yuan, Junjie Qiu, Junlong Li, Junxiao Song, Kai Dong, Kai Hu, Kaige Gao, Kang Guan, Kexin Huang, Kuai Yu, Lean Wang, Lecong Zhang, Lei Xu, Leyi Xia, Liang Zhao, Litong Wang, Liyue Zhang, Meng Li, Miaojun Wang, Mingchuan Zhang, Minghua Zhang, Minghui Tang, Mingming Li, Ning Tian, Panpan Huang, Peiyi Wang, Peng Zhang, Qiancheng Wang, Qihao Zhu, Qinyu Chen, Qiushi Du, R.~J. Chen, R.~L. Jin, Ruiqi Ge, Ruisong Zhang, Ruizhe Pan, Runji Wang, Runxin Xu, Ruoyu Zhang, Ruyi Chen, S.~S. Li, Shanghao Lu, Shangyan Zhou,
  Shanhuang Chen, Shaoqing Wu, Shengfeng Ye, Shengfeng Ye, Shirong Ma, Shiyu Wang, Shuang Zhou, Shuiping Yu, Shunfeng Zhou, Shuting Pan, T.~Wang, Tao Yun, Tian Pei, Tianyu Sun, W.~L. Xiao, and Wangding Zeng. 2024.
\newblock \href {https://doi.org/10.48550/ARXIV.2412.19437} {Deepseek-v3 technical report}.
\newblock \emph{CoRR}, abs/2412.19437.

\bibitem[{Dubey et~al.(2024)Dubey, Jauhri, Pandey, Kadian, Al{-}Dahle, Letman, Mathur, Schelten, Yang, Fan, Goyal, Hartshorn, Yang, Mitra, Sravankumar, Korenev, Hinsvark, Rao, Zhang, Rodriguez, Gregerson, Spataru, Rozi{\`{e}}re, Biron, Tang, Chern, Caucheteux, Nayak, Bi, Marra, McConnell, Keller, Touret, Wu, Wong, Ferrer, Nikolaidis, Allonsius, Song, Pintz, Livshits, Esiobu, Choudhary, Mahajan, Garcia{-}Olano, Perino, Hupkes, Lakomkin, AlBadawy, Lobanova, Dinan, Smith, Radenovic, Zhang, Synnaeve, Lee, Anderson, Nail, Mialon, Pang, Cucurell, Nguyen, Korevaar, Xu, Touvron, Zarov, Ibarra, Kloumann, Misra, Evtimov, Copet, Lee, Geffert, Vranes, Park, Mahadeokar, Shah, van~der Linde, Billock, Hong, Lee, Fu, Chi, Huang, Liu, Wang, Yu, Bitton, Spisak, Park, Rocca, Johnstun, Saxe, Jia, Alwala, Upasani, Plawiak, Li, Heafield, Stone, and et~al.}]{llama-3}
Abhimanyu Dubey, Abhinav Jauhri, Abhinav Pandey, Abhishek Kadian, Ahmad Al{-}Dahle, Aiesha Letman, Akhil Mathur, Alan Schelten, Amy Yang, Angela Fan, Anirudh Goyal, Anthony Hartshorn, Aobo Yang, Archi Mitra, Archie Sravankumar, Artem Korenev, Arthur Hinsvark, Arun Rao, Aston Zhang, Aur{\'{e}}lien Rodriguez, Austen Gregerson, Ava Spataru, Baptiste Rozi{\`{e}}re, Bethany Biron, Binh Tang, Bobbie Chern, Charlotte Caucheteux, Chaya Nayak, Chloe Bi, Chris Marra, Chris McConnell, Christian Keller, Christophe Touret, Chunyang Wu, Corinne Wong, Cristian~Canton Ferrer, Cyrus Nikolaidis, Damien Allonsius, Daniel Song, Danielle Pintz, Danny Livshits, David Esiobu, Dhruv Choudhary, Dhruv Mahajan, Diego Garcia{-}Olano, Diego Perino, Dieuwke Hupkes, Egor Lakomkin, Ehab AlBadawy, Elina Lobanova, Emily Dinan, Eric~Michael Smith, Filip Radenovic, Frank Zhang, Gabriel Synnaeve, Gabrielle Lee, Georgia~Lewis Anderson, Graeme Nail, Gr{\'{e}}goire Mialon, Guan Pang, Guillem Cucurell, Hailey Nguyen, Hannah Korevaar, Hu~Xu, Hugo
  Touvron, Iliyan Zarov, Imanol~Arrieta Ibarra, Isabel~M. Kloumann, Ishan Misra, Ivan Evtimov, Jade Copet, Jaewon Lee, Jan Geffert, Jana Vranes, Jason Park, Jay Mahadeokar, Jeet Shah, Jelmer van~der Linde, Jennifer Billock, Jenny Hong, Jenya Lee, Jeremy Fu, Jianfeng Chi, Jianyu Huang, Jiawen Liu, Jie Wang, Jiecao Yu, Joanna Bitton, Joe Spisak, Jongsoo Park, Joseph Rocca, Joshua Johnstun, Joshua Saxe, Junteng Jia, Kalyan~Vasuden Alwala, Kartikeya Upasani, Kate Plawiak, Ke~Li, Kenneth Heafield, Kevin Stone, and et~al. 2024.
\newblock \href {https://doi.org/10.48550/ARXIV.2407.21783} {The llama 3 herd of models}.
\newblock \emph{CoRR}, abs/2407.21783.

\bibitem[{Hooper et~al.(2024)Hooper, Kim, Mohammadzadeh, Mahoney, Shao, Keutzer, and Gholami}]{kvquant}
Coleman Hooper, Sehoon Kim, Hiva Mohammadzadeh, Michael~W. Mahoney, Yakun~Sophia Shao, Kurt Keutzer, and Amir Gholami. 2024.
\newblock \href {http://papers.nips.cc/paper\_files/paper/2024/hash/028fcbcf85435d39a40c4d61b42c99a4-Abstract-Conference.html} {Kvquant: Towards 10 million context length {LLM} inference with {KV} cache quantization}.
\newblock In \emph{Advances in Neural Information Processing Systems 38: Annual Conference on Neural Information Processing Systems 2024, NeurIPS 2024, Vancouver, BC, Canada, December 10 - 15, 2024}.

\bibitem[{Hsieh et~al.(2024)Hsieh, Sun, Kriman, Acharya, Rekesh, Jia, Zhang, and Ginsburg}]{ruler}
Cheng{-}Ping Hsieh, Simeng Sun, Samuel Kriman, Shantanu Acharya, Dima Rekesh, Fei Jia, Yang Zhang, and Boris Ginsburg. 2024.
\newblock \href {https://doi.org/10.48550/ARXIV.2404.06654} {{RULER:} what's the real context size of your long-context language models?}
\newblock \emph{CoRR}, abs/2404.06654.

\bibitem[{Jiang et~al.(2024)Jiang, Sablayrolles, Roux, Mensch, Savary, Bamford, Chaplot, de~Las~Casas, Hanna, Bressand, Lengyel, Bour, Lample, Lavaud, Saulnier, Lachaux, Stock, Subramanian, Yang, Antoniak, Scao, Gervet, Lavril, Wang, Lacroix, and Sayed}]{mixtral}
Albert~Q. Jiang, Alexandre Sablayrolles, Antoine Roux, Arthur Mensch, Blanche Savary, Chris Bamford, Devendra~Singh Chaplot, Diego de~Las~Casas, Emma~Bou Hanna, Florian Bressand, Gianna Lengyel, Guillaume Bour, Guillaume Lample, L{\'{e}}lio~Renard Lavaud, Lucile Saulnier, Marie{-}Anne Lachaux, Pierre Stock, Sandeep Subramanian, Sophia Yang, Szymon Antoniak, Teven~Le Scao, Th{\'{e}}ophile Gervet, Thibaut Lavril, Thomas Wang, Timoth{\'{e}}e Lacroix, and William~El Sayed. 2024.
\newblock \href {https://doi.org/10.48550/ARXIV.2401.04088} {Mixtral of experts}.
\newblock \emph{CoRR}, abs/2401.04088.

\bibitem[{Kim et~al.(2023)Kim, Hooper, Wattanawong, Kang, Yan, Genc, Dinh, Huang, Keutzer, Mahoney, Shao, and Gholami}]{transformer-survey}
Sehoon Kim, Coleman Hooper, Thanakul Wattanawong, Minwoo Kang, Ruohan Yan, Hasan Genc, Grace Dinh, Qijing Huang, Kurt Keutzer, Michael~W. Mahoney, Yakun~Sophia Shao, and Amir Gholami. 2023.
\newblock \href {https://doi.org/10.48550/ARXIV.2302.14017} {Full stack optimization of transformer inference: a survey}.
\newblock \emph{CoRR}, abs/2302.14017.

\bibitem[{Li et~al.(2024)Li, Huang, Yang, Venkitesh, Locatelli, Ye, Cai, Lewis, and Chen}]{snapkv}
Yuhong Li, Yingbing Huang, Bowen Yang, Bharat Venkitesh, Acyr Locatelli, Hanchen Ye, Tianle Cai, Patrick Lewis, and Deming Chen. 2024.
\newblock \href {http://papers.nips.cc/paper\_files/paper/2024/hash/28ab418242603e0f7323e54185d19bde-Abstract-Conference.html} {Snapkv: {LLM} knows what you are looking for before generation}.
\newblock In \emph{Advances in Neural Information Processing Systems 38: Annual Conference on Neural Information Processing Systems 2024, NeurIPS 2024, Vancouver, BC, Canada, December 10 - 15, 2024}.

\bibitem[{Linde et~al.(1980)Linde, Buzo, and Gray}]{lbg}
Yoseph Linde, Andres Buzo, and Robert~M. Gray. 1980.
\newblock \href {https://doi.org/10.1109/TCOM.1980.1094577} {An algorithm for vector quantizer design}.
\newblock \emph{{IEEE} Trans. Commun.}, 28(1):84--95.

\bibitem[{Lingle(2024)}]{transformer-vq}
Lucas~D. Lingle. 2024.
\newblock \href {https://openreview.net/forum?id=oDdzXQzP2F} {Transformer-vq: Linear-time transformers via vector quantization}.
\newblock In \emph{The Twelfth International Conference on Learning Representations, {ICLR} 2024, Vienna, Austria, May 7-11, 2024}. OpenReview.net.

\bibitem[{Liu et~al.(2024{\natexlab{a}})Liu, Li, Zhao, Zhang, and Guo}]{clusterkv}
Guangda Liu, Chengwei Li, Jieru Zhao, Chenqi Zhang, and Minyi Guo. 2024{\natexlab{a}}.
\newblock \href {https://doi.org/10.48550/ARXIV.2412.03213} {Clusterkv: Manipulating {LLM} {KV} cache in semantic space for recallable compression}.
\newblock \emph{CoRR}, abs/2412.03213.

\bibitem[{Liu et~al.(2024{\natexlab{b}})Liu, Yuan, Jin, Zhong, Xu, Braverman, Chen, and Hu}]{kivi}
Zirui Liu, Jiayi Yuan, Hongye Jin, Shaochen Zhong, Zhaozhuo Xu, Vladimir Braverman, Beidi Chen, and Xia Hu. 2024{\natexlab{b}}.
\newblock \href {https://openreview.net/forum?id=L057s2Rq8O} {{KIVI:} {A} tuning-free asymmetric 2bit quantization for {KV} cache}.
\newblock In \emph{Forty-first International Conference on Machine Learning, {ICML} 2024, Vienna, Austria, July 21-27, 2024}. OpenReview.net.

\bibitem[{OpenAI(2023)}]{gpt-4}
OpenAI. 2023.
\newblock \href {https://doi.org/10.48550/ARXIV.2303.08774} {{GPT-4} technical report}.
\newblock \emph{CoRR}, abs/2303.08774.

\bibitem[{Penedo et~al.(2024)Penedo, Kydl{\'{\i}}cek, Allal, Lozhkov, Mitchell, Raffel, von Werra, and Wolf}]{fineweb}
Guilherme Penedo, Hynek Kydl{\'{\i}}cek, Loubna~Ben Allal, Anton Lozhkov, Margaret Mitchell, Colin~A. Raffel, Leandro von Werra, and Thomas Wolf. 2024.
\newblock \href {http://papers.nips.cc/paper\_files/paper/2024/hash/370df50ccfdf8bde18f8f9c2d9151bda-Abstract-Datasets\_and\_Benchmarks\_Track.html} {The fineweb datasets: Decanting the web for the finest text data at scale}.
\newblock In \emph{Advances in Neural Information Processing Systems 38: Annual Conference on Neural Information Processing Systems 2024, NeurIPS 2024, Vancouver, BC, Canada, December 10 - 15, 2024}.

\bibitem[{Qin et~al.(2024)Qin, Li, He, Zhang, Wu, Zheng, and Xu}]{mooncake}
Ruoyu Qin, Zheming Li, Weiran He, Mingxing Zhang, Yongwei Wu, Weimin Zheng, and Xinran Xu. 2024.
\newblock \href {https://doi.org/10.48550/ARXIV.2407.00079} {Mooncake: {A} kvcache-centric disaggregated architecture for {LLM} serving}.
\newblock \emph{CoRR}, abs/2407.00079.

\bibitem[{Reid et~al.(2024)Reid, Savinov, Teplyashin, Lepikhin, Lillicrap, Alayrac, Soricut, Lazaridou, Firat, Schrittwieser, Antonoglou, Anil, Borgeaud, Dai, Millican, Dyer, Glaese, Sottiaux, Lee, Viola, Reynolds, Xu, Molloy, Chen, Isard, Barham, Hennigan, McIlroy, Johnson, Schalkwyk, Collins, Rutherford, Moreira, Ayoub, Goel, Meyer, Thornton, Yang, Michalewski, Abbas, Schucher, Anand, Ives, Keeling, Lenc, Haykal, Shakeri, Shyam, Chowdhery, Ring, Spencer, Sezener, and et~al.}]{gemini-1.5}
Machel Reid, Nikolay Savinov, Denis Teplyashin, Dmitry Lepikhin, Timothy~P. Lillicrap, Jean{-}Baptiste Alayrac, Radu Soricut, Angeliki Lazaridou, Orhan Firat, Julian Schrittwieser, Ioannis Antonoglou, Rohan Anil, Sebastian Borgeaud, Andrew~M. Dai, Katie Millican, Ethan Dyer, Mia Glaese, Thibault Sottiaux, Benjamin Lee, Fabio Viola, Malcolm Reynolds, Yuanzhong Xu, James Molloy, Jilin Chen, Michael Isard, Paul Barham, Tom Hennigan, Ross McIlroy, Melvin Johnson, Johan Schalkwyk, Eli Collins, Eliza Rutherford, Erica Moreira, Kareem Ayoub, Megha Goel, Clemens Meyer, Gregory Thornton, Zhen Yang, Henryk Michalewski, Zaheer Abbas, Nathan Schucher, Ankesh Anand, Richard Ives, James Keeling, Karel Lenc, Salem Haykal, Siamak Shakeri, Pranav Shyam, Aakanksha Chowdhery, Roman Ring, Stephen Spencer, Eren Sezener, and et~al. 2024.
\newblock \href {https://doi.org/10.48550/ARXIV.2403.05530} {Gemini 1.5: Unlocking multimodal understanding across millions of tokens of context}.
\newblock \emph{CoRR}, abs/2403.05530.

\bibitem[{Singhania et~al.(2024)Singhania, Singh, He, Feizi, and Bhatele}]{loki}
Prajwal Singhania, Siddharth Singh, Shwai He, Soheil Feizi, and Abhinav Bhatele. 2024.
\newblock \href {http://papers.nips.cc/paper\_files/paper/2024/hash/1e027da6bec9ceb2ec37951ceeccae93-Abstract-Conference.html} {Loki: Low-rank keys for efficient sparse attention}.
\newblock In \emph{Advances in Neural Information Processing Systems 38: Annual Conference on Neural Information Processing Systems 2024, NeurIPS 2024, Vancouver, BC, Canada, December 10 - 15, 2024}.

\bibitem[{Su(2023{\natexlab{a}})}]{rerope-blog}
Jianlin Su. 2023{\natexlab{a}}.
\newblock Expand the context length with rope, part 3 -- unlocking the unlimited extrapolation potential with rerope.
\newblock \url{https://normxu.github.io/Rethinking-Rotary-Position-Embedding-3/}.

\bibitem[{Su(2023{\natexlab{b}})}]{rerope}
Jianlin Su. 2023{\natexlab{b}}.
\newblock Rectified rotary position embeddings.
\newblock \url{https://github.com/bojone/rerope}.

\bibitem[{Su et~al.(2021)Su, Lu, Pan, Wen, and Liu}]{rope}
Jianlin Su, Yu~Lu, Shengfeng Pan, Bo~Wen, and Yunfeng Liu. 2021.
\newblock \href {https://arxiv.org/abs/2104.09864} {Roformer: Enhanced transformer with rotary position embedding}.
\newblock \emph{CoRR}, abs/2104.09864.

\bibitem[{Tang et~al.(2024)Tang, Zhao, Zhu, Xiao, Kasikci, and Han}]{quest}
Jiaming Tang, Yilong Zhao, Kan Zhu, Guangxuan Xiao, Baris Kasikci, and Song Han. 2024.
\newblock \href {https://openreview.net/forum?id=KzACYw0MTV} {{QUEST:} query-aware sparsity for efficient long-context {LLM} inference}.
\newblock In \emph{Forty-first International Conference on Machine Learning, {ICML} 2024, Vienna, Austria, July 21-27, 2024}. OpenReview.net.

\bibitem[{Vaswani et~al.(2017)Vaswani, Shazeer, Parmar, Uszkoreit, Jones, Gomez, Kaiser, and Polosukhin}]{transformer}
Ashish Vaswani, Noam Shazeer, Niki Parmar, Jakob Uszkoreit, Llion Jones, Aidan~N. Gomez, Lukasz Kaiser, and Illia Polosukhin. 2017.
\newblock \href {https://arxiv.org/abs/1706.03762} {Attention is all you need}.
\newblock \emph{CoRR}, abs/1706.03762.

\bibitem[{Wu et~al.(2024)Wu, Song, and Duthie}]{megabeam-mistral}
Chen Wu, Yin Song, and Eden Duthie. 2024.
\newblock \href {https://huggingface.co/aws-prototyping/MegaBeam-Mistral-7B-512k} {{aws-prototyping/MegaBeam-Mistral-7B-512k}}.

\bibitem[{Xiao et~al.(2024)Xiao, Tian, Chen, Han, and Lewis}]{streaming-llm}
Guangxuan Xiao, Yuandong Tian, Beidi Chen, Song Han, and Mike Lewis. 2024.
\newblock \href {https://openreview.net/forum?id=NG7sS51zVF} {Efficient streaming language models with attention sinks}.
\newblock In \emph{The Twelfth International Conference on Learning Representations, {ICLR} 2024, Vienna, Austria, May 7-11, 2024}. OpenReview.net.

\bibitem[{Yang et~al.(2024{\natexlab{a}})Yang, Yang, Zhang, Hui, Zheng, Yu, Li, Liu, Huang, Wei, Lin, Yang, Tu, Zhang, Yang, Yang, Zhou, Lin, Dang, Lu, Bao, Yang, Yu, Li, Xue, Zhang, Zhu, Men, Lin, Li, Xia, Ren, Ren, Fan, Su, Zhang, Wan, Liu, Cui, Zhang, and Qiu}]{qwen-2.5}
An~Yang, Baosong Yang, Beichen Zhang, Binyuan Hui, Bo~Zheng, Bowen Yu, Chengyuan Li, Dayiheng Liu, Fei Huang, Haoran Wei, Huan Lin, Jian Yang, Jianhong Tu, Jianwei Zhang, Jianxin Yang, Jiaxi Yang, Jingren Zhou, Junyang Lin, Kai Dang, Keming Lu, Keqin Bao, Kexin Yang, Le~Yu, Mei Li, Mingfeng Xue, Pei Zhang, Qin Zhu, Rui Men, Runji Lin, Tianhao Li, Tingyu Xia, Xingzhang Ren, Xuancheng Ren, Yang Fan, Yang Su, Yichang Zhang, Yu~Wan, Yuqiong Liu, Zeyu Cui, Zhenru Zhang, and Zihan Qiu. 2024{\natexlab{a}}.
\newblock \href {https://doi.org/10.48550/ARXIV.2412.15115} {Qwen2.5 technical report}.
\newblock \emph{CoRR}, abs/2412.15115.

\bibitem[{Yang et~al.(2024{\natexlab{b}})Yang, Han, Gao, Hu, Zhang, and Zhao}]{pyramidinfer}
Dongjie Yang, Xiaodong Han, Yan Gao, Yao Hu, Shilin Zhang, and Hai Zhao. 2024{\natexlab{b}}.
\newblock \href {https://doi.org/10.18653/V1/2024.FINDINGS-ACL.195} {Pyramidinfer: Pyramid {KV} cache compression for high-throughput {LLM} inference}.
\newblock In \emph{Findings of the Association for Computational Linguistics, {ACL} 2024, Bangkok, Thailand and virtual meeting, August 11-16, 2024}, pages 3258--3270. Association for Computational Linguistics.

\bibitem[{Zhang et~al.(2024)Zhang, Ji, Chen, Fu, Miao, Nie, Chen, and Cui}]{pqcache}
Hailin Zhang, Xiaodong Ji, Yilin Chen, Fangcheng Fu, Xupeng Miao, Xiaonan Nie, Weipeng Chen, and Bin Cui. 2024.
\newblock \href {https://doi.org/10.48550/ARXIV.2407.12820} {Pqcache: Product quantization-based kvcache for long context {LLM} inference}.
\newblock \emph{CoRR}, abs/2407.12820.

\bibitem[{Zhang et~al.(2023)Zhang, Sheng, Zhou, Chen, Zheng, Cai, Song, Tian, R{\'{e}}, Barrett, Wang, and Chen}]{h2o}
Zhenyu Zhang, Ying Sheng, Tianyi Zhou, Tianlong Chen, Lianmin Zheng, Ruisi Cai, Zhao Song, Yuandong Tian, Christopher R{\'{e}}, Clark~W. Barrett, Zhangyang Wang, and Beidi Chen. 2023.
\newblock \href {http://papers.nips.cc/paper\_files/paper/2023/hash/6ceefa7b15572587b78ecfcebb2827f8-Abstract-Conference.html} {{H2O:} heavy-hitter oracle for efficient generative inference of large language models}.
\newblock In \emph{Advances in Neural Information Processing Systems 36: Annual Conference on Neural Information Processing Systems 2023, NeurIPS 2023, New Orleans, LA, USA, December 10 - 16, 2023}.

\end{thebibliography}
